# Effects of Discretization of Decision and Objective Spaces on the Performance of Evolutionary Multi-objective Optimization Algorithms

Weiyu Chen, Hisao Ishibuchi, Ke Shang

Shenzhen Key Laboratory of Computational Intelligence,
University Key Laboratory of Evolving Intelligent Systems of Guangdong Province,
Department of Computer Science and Engineering, Southern University of Science and Technology (SUSTech), Shenzhen, China
Email: 11711904@mail.sustech.edu.cn; hisao@sustech.edu.cn; kshang@foxmail.com

*Abstract*—Recently, the discretization of decision and objective spaces has been discussed in the literature. In some studies, it is shown that the decision space discretization improves the performance of evolutionary multi-objective optimization (EMO) algorithms on continuous multi-objective test problems. In other studies, it is shown that the objective space discretization improves the performance on combinatorial multi-objective problems. However, the effect of the simultaneous discretization of both spaces has not been examined in the literature. In this paper, we examine the effects of the decision space discretization, objective space discretization and simultaneous discretization on the performance of NSGA-II through computational experiments on the DTLZ and WFG problems. Using various settings about the number of decision variables and the number of objectives, our experiments are performed on four types of problems: standard problems, large-scale problems, many-objective problems, and large-scale many-objective problems. We show that the decision space discretization has a positive effect for large-scale problems and the objective space discretization has a positive effect for many-objective problems. We also show the discretization of both spaces is useful for large-scale many-objective problems.

*Keywords—evolutionary multi-objective optimizati，multi-objective problem, many-objective problem，large-scale problems，decision space discretizatio，objective space discretization.*

## I. INTRODUCTION

Over the past few decades, a number of evolutionary multi-objective optimization (EMO) algorithms have been proposed. They are often categorized into three classes: (i) Pareto dominance-based algorithms (e.g., NSGA-II [1], SPEA2 [2]), (ii) Decomposition-based algorithms (e.g., MOEA/D [3]) and (iii) Indicator-based algorithms (e.g., SMS-EMOA [4]). Whereas Pareto dominance-based algorithms have been frequently used, their search ability is not always high for many-objective problems and large-scale problems.

A simple but potentially useful idea for improving EMO algorithms is the discretization of decision variables and/or objective functions. Kondo et al. [5] examined the effect of the decision space discretization on the performance of NSGA-II [1]. They reported that the coarse discretization can accelerate the convergence but it worsens the diversity. Based on their observations, they proposed an idea of adaptive discretization of the decision space. The resolution is chosen adaptively based on a standard deviation (SD) indicator or an estimated probability density function. They showed that the proposed idea can improve the performance of NSGA-II [1] on continuous multi-objective test problems. Ishibushi et al. [6], [7] examined the objective space discretization. They reported that the objective discretization improves the search ability of NSGA-II [1] and SPEA2 [2] on combinatorial many-objective problems.

However, the effect of the simultaneous discretization of both spaces has not been examined in the literature. In this paper, we examine the performance of NSGA-II [1] on the frequently-used DTLZ [8] and WFG [9] test problems under the following four settings of the discretization: no discretization, decision space discretization, objective space discretization, and simultaneous discretization of both spaces. To examine the performance of NSGA-II not only on multi-objective problems but also on many-objective and large-scale problems. We use a wide range of specifications for the number of objectives (3, 5, 10, 15 objectives) and the number of decision variables (20 and 1000 decision variables).

This paper is organized as follows. In Section II, we explain our experimental settings. In Section III, we explain our decision space discretization method and report its effects on the performance of NSGA-II on our test problems. In Section IV, we explain our objective space discretization method and report its effects. In Section V, we report our experimental results based on the simultaneous discretization of both spaces. Finally we draw a conlcusion in Section VI.

## II. EXPERIMENTAL SETTINGS

To examine the effect of the discretization on standard multi-objective problems, large-scale multi-objective problems, many-objective problems, and large-scale many-objective problems, we use 10 benchmark problems: 5 DTLZ problems [8] and 5 WFG problems [9]. Table I summarizes the properties of

This work was supported by National Natural Science Foundation of China (Grant No. 61876075), the Program for Guangdong Introducing Innovative and Entrepreneurial Teams (Grant No. 2017ZT07X386), Shenzhen Peacock Plan (Grant No. KQTD2016112514355531), the Science and Technology Innovation Committee Foundation of Shenzhen (Grant No. ZDSYS201703031748284), and the Program for University Key Laboratory of Guangdong Province (Grant No. 2017KSYS008).

Corresponding Author: Hisao Ishibuchi (hisao@sustech.edu.cn)

TABLE I. PROPERTIES OF BENCHMARK PROBLEMS [9]

| Problems | Separability | Modality | Bias | Pareto front Geometry |
|---|---|---|---|---|
| DTLZ1 | Separable | M | - | Linear |
| DTLZ2 | Separable | U | - | Concave |
| DTLZ3 | Separable | M | - | Concave |
| DTLZ4 | Separable | U | + | Concave |
| DTLZ5 | ? | U | - | ? |
| WFG1 | Separable | U | + | Convex, Mixed |
| WFG2 | Non-separable | M | - | Convex, Disconnected |
| WFG3 | Non-separable | U | - | Linear, Degenerate (1) |
| WFG4 | Separable | M | - | Concave |
| WFG5 | Separable | D | - | Concave |

(1) Recently it was reported the Pareto front of WFG3 is not degenerate [16].

TABLE II. PARAMETER SETTINGS

| Common Parameters | Population Size | 100 | | |
| | Maximum Generations | 200 | | |
| | Crossover rate | 1.0 | | |
| | Mutation rate | 1/ n | | |
| Standard Multi-objective Problem | Number of Objectives | 3 | | |
| | Number of Decision Variables | 7 | | |
| Large-scale Multi-objective Problem | Number of Objectives | 3 | | |
| | Number of Decision Variables | 1000 | | |
| Many-objective Problem | Number of Objectives | 5 | 10 | 15 |
| | Number of Decision Variables | 9 | 14 | 19 |
| Large-scale many-objective Problem | Number of Objectives | 5 | 10 | 15 |
| | Number of Decision Variables | 1000 | | |

these problems (which is cited from [9]). In the third column 'M' means Multimodal, 'U' means Unimodal and 'D' means Deceptive. In the fourth column, '+' and '-' are used to indicate biased and unbiased problems, respectively. We use these test problems since they have been almost always used in the EMO community to evaluate the performance of EMO algorithms on multi-objective and many-objective problems (whereas these test problems do not have a variety of Pareto fronts in the sense that most of them have triangular Pareto fronts [17]).

NSGA-II with polynomial mutation and simulated binary crossover (SBX) [10] is used in our computational experiments. Parameter values in NSGA-II are shown in Table II. The algorithm is terminated after 200 generations. Table II also shows the specifications of the number of objectives and the number of decision variables in the following four types of test problems: standard multi-objective problems, large-scale multi-objective problems, many-objective problems, and large-scale many-objective problems. As shown in Table II, "large-scale" and "many-objective" are characterized by 1000 decision variables and 5-15 objectives, respectively, in our experiments.

We choose the inverted generational distance (IGD) metric [11], [12] as the performance indicator, which can be used to evaluate both the convergence and the diversity. We use the given reference points in PlatEMO [13] for IGD calculation.

TABLE III. AVERAGE IGD VALUES BY NSGA-II AND NSGA-II-DD ON THE LARGE-SCALE THREE-OBJECTIVE PROBLEMS WITH 1000 DECISION VARIABLES. THE BLUE BOLD FONT SSOWS THE BETTER RESULT

| Problems | NSGA-II | NSGA-II-DD |
|---|---|---|
| DTLZ1 | 2.3397e+4 (5.46e+2) | **2.0562e+4 (5.00e+2) +** |
| DTLZ2 | 2.3120e+1 (9.83e-1) | **2.1591e+1 (9.77e-1) +** |
| DTLZ3 | 7.2109e+4 (1.30e+3) | **6.1626e+4 (1.80e+3) +** |
| DTLZ4 | 2.4532e+1 (2.96e+0) | **2.2365e+1 (1.53e+0) +** |
| DTLZ5 | 2.6346e+1 (1.23e+0) | **2.5484e+1 (1.35e+0) +** |
| WFG1 | **1.5184e+0 (1.11e-2)** | 2.2653e+0 (4.42e-2) - |
| WFG2 | 6.3614e-1 (3.48e-2) | **5.7797e-1 (1.34e-2) +** |
| WFG3 | 7.4098e-1 (1.71e-2) | **7.2231e-1 (1.97e-2) +** |
| WFG4 | 4.6906e-1 (9.43e-3) | **4.3047e-1 (9.24e-3) +** |
| WFG5 | 6.2287e-1 (8.58e-3) | **5.8818e-1 (8.01e-3) +** |
| +/-/= | | 9 / 1 / 0 |

In our computational experiments, we use the following four versions of NSGA-II to examine the effect of the discretization:

1. **NSGA-II**: The original NSGA-II with no discretization.
2. **NSGA-II-DD**: NSGA-II with Decision space Discretization.
3. **NSGA-II-OD**: NSGA-II with Objective space Discretization.
4. **NSGA-II-BD**: NSGA-II with Both-space Discretization.

Each NSGA-II algorithm is executed 30 times for each test problem. The average IGD values and standard deviation are calculated for each algorithm and each test problem (as shown in Tables III-VII, which are explained in detail later in this paper). Experimental results are analyzed by the Wilcoxon rank sum test with a significance level of 0.05 to check whether one algorithm obtains a statistically significant better result than another algorithm. The test results are shown by "+", "–" and "=" in Tables III-VII where "+", "–" and "=" indicate that each algorithm is "significantly better than", "significantly worse than" and "not significantly different from" the original NSGA-II algorithm, respectively. For each table, the best average IGD value for each problem is highlighted by a blue bold font.

III. DISCRETIZATION IN DECISION SPACE

A. Discretization Method

We use the discretization method of Kondo et al. [5]. The resolution of discretization is chosen adaptively according to the standard deviation (SD) indicator. The number of decimal places $d_i$ for the $i$th decision variable $x_i$ is specified by the following formula:

$$d_i = round\left(\left(1 - \frac{\sigma_i}{\sigma_{max}}\right)(d_{max} - d_{min}) + d_{min}\right)$$

where $\sigma_i$ is the standard deviation of $x_i$. $\sigma_{max}$ is assumed as the standard deviation of the uniform distribution in [0,1], which is $1/\sqrt{12}$. The values of $d_{min}$ and $d_{max}$ in our experiment are 2 and 8, respectively. For more explanations, see [5].

B. Results on Large-Scale Problems

The results of NSGA-II and NSGA-II-DD for the large-scale problems (see Table II) are compared in Table III. Table III

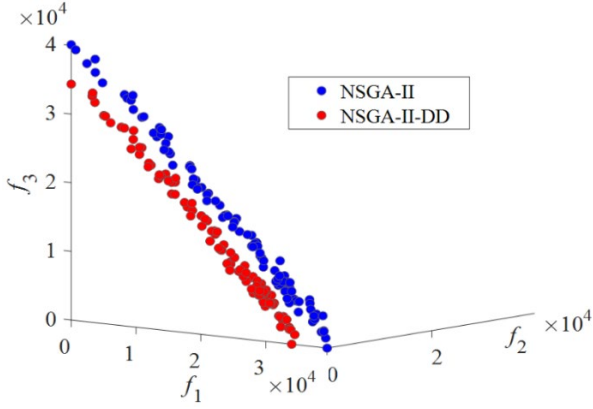

**Fig. 1.** Non-dominated solutions obtained by NSGA-II and NSGA-II-DD on large-scale DTLZ1 problem.

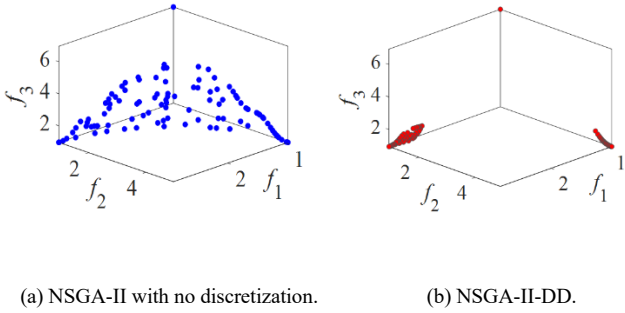

(a) NSGA-II with no discretization.  (b) NSGA-II-DD.

**Fig. 2.** Obtained non-dominated solutions by a single run of NSGA-II with no discretization and NSGA-II-DD with the decision space discretization.

shows that the decision space discretization has a clear positive effect. NSGA-II-DD outperforms NSGA-II for almost all test problems except for WFG1 as shown by the blue bold font.

Our results in Table I are consistent with the reported results in Kondo et al. [5] That is, the decision space discretization improved the performance of NSGA-II. The reason for the performance improvement on the large-scale problems can be explained as follows. The decision space discretization decreases the size of the search space by decreasing the number of possible solutions. By reducing the resolution of the discretization, we can decrease the number of possible solutions. The effect of the search space reduction has a positive effect on the ability of search algorithms when the search space is huge as in the large-scale problem. To further discuss this issue, we show the non-dominated solution set obtained by a single run of NSGA-II and NSGA-II-DD on the large-scale three-objective DTLZ1 problem in Fig. 1. We choose a single run whose IGD value is closest to the average IGD value among the 30 runs. We can see from Fig. 1 that the convergence ability of NSGA-II is improved by the decision space discretization in NSGA-II-DD.

However, the decision space discretization also has a negative effect on the diversity of solutions. In Fig. 2, we show the obtained non-dominated solutions by a single run of each algorithm on the large-scale three-objective WFG1 problem.

TABLE IV. AVERAGE IGD VALUES BY NSGA-II AND NSGA-II-DD ON MANY-OBJECTIVE PROBLEMS. THE BLUE BOLD FONT SHOWS THE BETTER RESULT

| Problem | M | NSGA-II | NSGA-II-DD |
|---|---|---|---|
| DTLZ1 | 5 | 9.7495e-1 (7.51e-1) | **7.4840e-1 (7.35e-1)** = |
|  | 10 | 2.0814e+1 (7.63e+0) | **1.8381e+1 (7.86e+0)** = |
|  | 15 | 2.4219e+1 (1.03e+1) | **2.0848e+1 (8.88e+0)** = |
| DTLZ2 | 5 | 2.6123e-1 (8.36e-3) | **2.5239e-1 (6.58e-3)** + |
|  | 10 | 1.7860e+0 (3.73e-1) | **1.7309e+0 (3.31e-1)** = |
|  | 15 | **1.4866e+0 (1.28e-1)** | 1.5037e+0 (1.16e-1) = |
| DTLZ3 | 5 | 2.6552e+1 (1.62e+1) | **6.1674e+0 (3.13e+0)** + |
|  | 10 | **1.1700e+3 (2.61e+2)** | 1.1959e+3 (2.56e+2) = |
|  | 15 | **9.2318e+2 (2.19e+2)** | 9.3412e+2 (2.79e+2) = |
| DTLZ4 | 5 | **2.5704e-1 (8.95e-3)** | 2.8448e-1 (1.08e-2) - |
|  | 10 | 1.5965e+0 (1.86e-1) | **1.2340e+0 (1.23e-1)** + |
|  | 15 | 1.5177e+0 (1.09e-1) | **1.4500e+0 (9.97e-2)** + |
| DTLZ5 | 5 | 1.1331e-1 (3.39e-2) | **1.0679e-1 (2.65e-2)** = |
|  | 10 | 4.8161e-1 (2.66e-1) | **4.5656e-1 (2.23e-1)** = |
|  | 15 | **6.7112e-1 (2.14e-1)** | 7.4999e-1 (2.14e-1) = |
| WFG1 | 5 | **1.4794e+0 (1.05e-1)** | 1.7578e+0 (2.00e-1) - |
|  | 10 | 2.6710e+0 (1.25e-1) | **2.2506e+0 (5.45e-2)** + |
|  | 15 | 3.4464e+0 (1.81e-1) | **2.9450e+0 (1.16e-1)** + |
| WFG2 | 5 | 8.2094e-1 (8.69e-2) | **7.9053e-1 (1.25e-1)** = |
|  | 10 | 2.2142e+0 (4.08e-1) | **2.1570e+0 (3.50e-1)** = |
|  | 15 | **1.2171e+0 (8.81e-1)** | 1.3656e+0 (1.06e+0) = |
| WFG3 | 5 | 5.4385e-1 (1.05e-1) | **5.1262e-1 (8.26e-2)** = |
|  | 10 | 2.1693e+0 (3.13e-1) | **1.9801e+0 (3.23e-1)** + |
|  | 15 | **3.8796e+0 (7.89e-1)** | 4.0212e+0 (7.94e-1) = |
| WFG4 | 5 | **1.2952e+0 (2.70e-2)** | 1.3152e+0 (2.49e-2) - |
|  | 10 | **5.1535e+0 (5.85e-2)** | 5.2050e+0 (7.39e-2) - |
|  | 15 | **9.1279e+0 (1.01e-1)** | 9.1548e+0 (1.05e-1) = |
| WFG5 | 5 | **1.2744e+0 (2.77e-2)** | 1.2771e+0 (2.14e-2) = |
|  | 10 | 5.1465e+0 (5.24e-2) | **5.1140e+0 (6.86e-2)** + |
|  | 15 | 9.1136e+0 (1.02e-1) | **8.9987e+0 (9.53e-2)** + |
| + / - / = |  |  | 9 / 4 / 17 |

*C. Results on Many-Objective Problems*

Let us examine the effect of the decision space discretization for many-objective problems. Table IV shows experimental results. In this table, the decision space discretization has a positive effect on some problems and a negative effect on other problems. Discretization of decision space does not have a positive effect on most problems. NSGA-II-DD outperforms NSGA-II on 9 problems and is outperformed by NSGA-II on 4 problems.

One interesting observation in Table IV is that NSGA-II-DD obtains better results than NSGA-II on the 10-objective and 15-objective WFG1 problems (whereas NSGA-II is better for the 3-objective and 5-objective WFG1 problems). As shown in Fig. 2, the decision space discretization worsens the diversity of solutions. However, better results are obtained by NSGA-II-DD on WFG1 with 10 and 15 objectives in Table IV. To further examine this observation, we show obtained non-dominated solutions by a single run of NSGA-II and NSGA-II-DD on the 10-objective WFG1 problem in Fig. 3. We can observe a negative effect of the decision space discretization on the

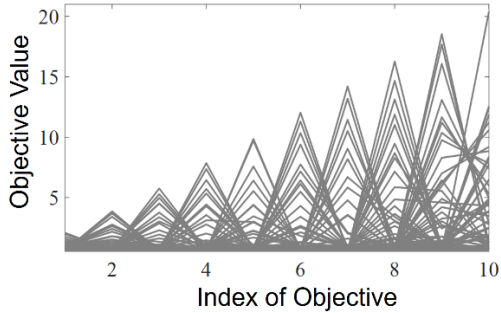
(a) NSGA-II with no discretization.

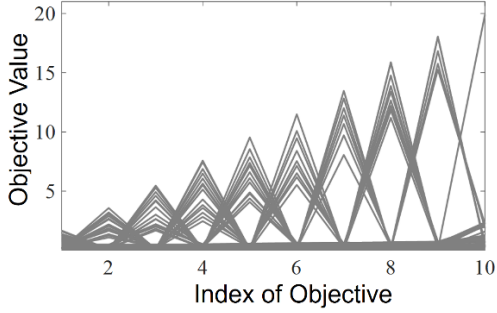
(b) NSGA-II-DD with the decision space discretization.

**Fig. 3.** Obtained non-dominated solutions by a single run of NSGA-II and NSGA-II-DD on the 10-objective WFG1 problem.

diversity of solutions in Fig. 3. However, NSGA-II-DD is better for WFG1 with 10 objectives in Table IV. To further discuss this apparent inconsistency, we calculate the generational distance (GD) value for each solution set in Fig. 3. The GD value is 0.24735 in Fig. 3 (a) by NSGA-II and 0.1414 in Fig. 3 (b). These GD values suggest that the better IGD values by NSGA-II-DD for WFG1 with 10 and 15 objectives in Table IV are due to the convergence ability improvement.

## IV. DISCRETIZATION IN OBJECTIVE SPACE

### A. Discretization Method

In this paper, the objective space discretization is performed as follows. Before the non-dominated sorting in NSGA-II, we normalize the objective values into [0, 1], and reserve two decimal places of the normalized objective values. For example, if we have an objective vector $f(x) = (450, 523, 651, 733, 869)$, first we normalize it to $f'(x) = (0.450, 0.523, 0.651, 0.733, 0.869)$. Then, we apply 2-digits discretization and get $f''(x) = (0.45, 0.52, 0.65, 0.73, 0.87)$. Then, $f''(x)$ is used for non-dominated sorting. This is a bit different from the epsilon dominance approach proposed in several papers [14], [15], but both methods have similar effects (i.e., they reduce the number of non-dominated solutions).

### B. Results on Large-Scale Problems

Experimental results obtained by NSGA-II and NSGA-II-OD on the large-scale three-objective problems with 1000 decision variables are shown in Table V. Whereas the decision space discretization improved the performance of NSGA-II on almost all problems in Table III (9 out of the 10 problems), the objective

TABLE V. AVERAGE IGD VALUES BY NSGA-II AND NSGA-II-OD ON THE LARGE-SCALE THREE-OBJECTIVE PROBLEMS WITH 1000 DECISION VARIABLES. THE BLUE BOLD FONT SHOWS THE BETTER RESULT

| Problems | NSGA-II | NSGA-II-OD |
|---|---|---|
| DTLZ1 | 2.3397e+4 (5.46e+2) | **2.1755e+4 (6.35e+2)** + |
| DTLZ2 | 2.3120e+1 (9.83e-1) | **1.9622e+1 (1.10e+0)** + |
| DTLZ3 | 7.2109e+4 (1.30e+3) | **6.5461e+4 (2.16e+3)** + |
| DTLZ4 | 2.4532e+1 (2.96e+0) | **2.2311e+1 (1.59e+0)** + |
| DTLZ5 | 2.6346e+1 (1.23e+0) | **1.9499e+1 (1.09e+0)** + |
| WFG1 | **1.5184e+0 (1.11e-2)** | 1.5337e+0 (1.34e-2) - |
| WFG2 | 6.3614e-1 (3.48e-2) | **6.1793e-1 (2.54e-2)** + |
| WFG3 | 7.4098e-1 (1.71e-2) | **6.7081e-1 (1.53e-2)** + |
| WFG4 | **4.6906e-1 (9.43e-3)** | 5.2148e-1 (8.24e-3) - |
| WFG5 | **6.2287e-1 (8.58e-3)** | 6.7227e-1 (1.36e-2) - |
| + / - / = | | **7 / 3 / 0** |

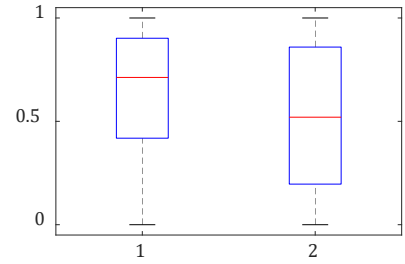
(a) NSGA-II with no discretization.

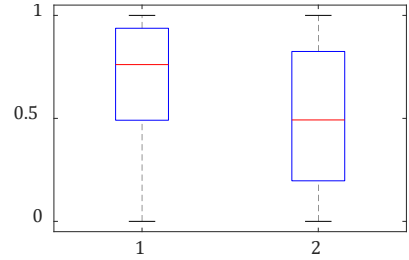
(b) NSGA-II-OD with the objective space discretization.

**Fig. 4.** Distribution of values of the first two decision variables in the non-dominated solution set obtained by a single run of NSGA-II and NSGA-II-OD on the large-scale three-objective DTLZ1 problem.

space discretization does not always improve the performance of NSGA-II in Table V (7 out of the 10 problems). However, from close comparison between Table III and Table V, we can see that better results are obtained from NSGA-II-OD in Table V than NSGA-II-DD in Table III for DTLZ2, DTLZ4, DTLZ5, WFG1 and WFG3. This is an interesting observation since all of these test problems are unimodal problems as shown in Table I. That is, Table III and Table V suggest that the objective space discretization is useful for unimodal problems while the decision space discretization is useful for multimodal problems.

To further discuss the experimental results in Table V, we focus on a single run of each algorithm on DTLZ1. As in the previous section, we select a single run whose IGD value is the closest to the average IGD value over 30 runs of each algorithm.

TABLE VI. AVERAGE IGD VALUES BY NSGA-II AND NSGA-II-OD ON MANY-OBJECTIVE PROBLEMS. THE BLUE BOLD FONT SHOWS THE BETTER RESULT

| Problems | M | NSGA-II | NSGA-II-OD |
|---|---|---|---|
| DTLZ1 | 5 | **1.0358e+0 (8.53e-1)** | 1.0914e+0 (4.63e-1) = |
| | 10 | 1.9347e+1 (8.10e+0) | **4.1015e+0 (6.86e+0) +** |
| | 15 | 2.4810e+1 (8.25e+0) | **1.6540e+1 (1.91e+1) +** |
| DTLZ2 | 5 | 2.6589e-1 (1.18e-2) | **2.3010e-1 (5.29e-3) +** |
| | 10 | 1.7084e+0 (3.31e-1) | **1.2062e+0 (1.21e-1) +** |
| | 15 | 1.4559e+0 (1.75e-1) | **1.3329e+0 (1.07e-1) +** |
| DTLZ3 | 5 | 2.4563e+1 (1.31e+1) | **3.5015e+0 (1.24e+0) +** |
| | 10 | 1.1666e+3 (2.48e+2) | **1.8968e+2 (6.08e+1) +** |
| | 15 | 8.6313e+2 (2.73e+2) | **2.1683e+2 (8.03e+1) +** |
| DTLZ4 | 5 | 2.5678e-1 (1.08e-2) | **2.5166e-1 (9.16e-2) +** |
| | 10 | 1.5783e+0 (1.53e-1) | **5.8944e-1 (1.37e-2) +** |
| | 15 | 1.5782e+0 (1.30e-1) | **7.4646e-1 (3.48e-2) +** |
| DTLZ5 | 5 | 1.0470e-1 (2.95e-2) | **8.4782e-2 (2.35e-2) +** |
| | 10 | 4.7673e-1 (1.79e-1) | **2.9855e-1 (8.33e-2) +** |
| | 15 | 7.4791e-1 (2.22e-1) | **3.7075e-1 (1.43e-1) +** |
| WFG1 | 5 | 1.5357e+0 (1.28e-1) | **1.3630e+0 (1.24e-1) +** |
| | 10 | 2.6599e+0 (1.25e-1) | **2.4796e+0 (1.61e-1) +** |
| | 15 | 3.4526e+0 (1.82e-1) | **3.1077e+0 (1.43e-1) +** |
| WFG2 | 5 | 8.4829e-1 (7.98e-2) | **7.0630e-1 (8.79e-2) +** |
| | 10 | **2.3969e+0 (4.35e-1)** | 2.4903e+0 (3.59e-1) = |
| | 15 | **1.3727e+0 (9.79e-1)** | 3.1687e+0 (9.00e-1) - |
| WFG3 | 5 | 5.2618e-1 (1.05e-1) | **4.2395e-1 (5.71e-2) +** |
| | 10 | 1.9765e+0 (4.07e-1) | **1.7669e+0 (2.85e-1) =** |
| | 15 | 3.8537e+0 (7.78e-1) | **3.6130e+0 (6.49e-1) =** |
| WFG4 | 5 | 1.2928e+0 (1.92e-2) | **1.2857e+0 (2.18e-2) =** |
| | 10 | 5.1485e+0 (7.62e-2) | **5.1056e+0 (6.40e-2) +** |
| | 15 | **9.1142e+0 (1.09e-1)** | 9.1231e+0 (8.10e-2) = |
| WFG5 | 5 | 1.2729e+0 (1.88e-2) | **1.2686e+0 (2.66e-2) =** |
| | 10 | 5.1669e+0 (7.65e-2) | **5.1003e+0 (7.31e-2) +** |
| | 15 | 9.0933e+0 (1.03e-1) | **9.0679e+0 (1.06e-1) =** |
| + / - / = | | | 21 / 1 / 8 |

Fig. 4 shows the distribution of values of each of the first two decision variables (i.e., $x_1$ and $x_2$) in the solutions obtained by the selected single run in a boxplot form. On the horizontal axis, "1" and "2" mean $x_1$ and $x_2$, respectively. The vertical axis is the value of each decision variable. As expected, the objective space discretization slightly decreases the diversity of solutions. This is because different solutions in the decision space can be viewed as the same solution in the objective space due to the objective space discretization. This may increase the difficulty of multimodal problems. Further analysis is needed as a future study to discussion the relation between the multimodality of problems and the usefulness of the decision space and/or objective space discretization.

## C. Results on Many-Objective Problems

Experimental results on the many-objective problems are shown in Table VI. It is clearly demonstrated that the objective space discretization has positive effects on the performance of NSGA-II on the many-objective problems. As shown in Table VI, NAGA-II-OD obtained better results than NSGA-II in 21 out of the 30 problems. These results are consistent with the results on combinatorial many-objective problems in [6].

As repeatedly reported in the literature, many-objective problems are not easy for dominance-based EMO algorithms. The reason is that almost all solutions with many objectives in the population are non-dominated. As a result, no strong selection pressure can be generated by the Pareto dominance relation. The objective space discretization can improve the search ability by decreasing the number of non-dominated solutions in the population. For example, assume that we have the following two non-dominated solutions for a five-objective minimization problem: (450, 523, 651, 733, 869) and (453, 510, 621, 703, 870) in the objective space. After the objective space discretization, they are handled as (0.45, 0.52, 0.65, 0.73, 0.87) and (0.45, 0.51, 0.62, 0.70, 0.87) where the first solution is dominated by the second one.

## V. DISCRETIZATION IN DECISION AND OBJECTIVE SPACES

From the above-mentioned experimental results, we can see that the decision space discretization (DD) and the objective space discretization (OD) have positive effects on the search ability of NSGA-II for large-scale multi-objective problems and many-objective problems, respectively. The main research question in this paper is whether the use of discretization in both of the decision and objective spaces further improves the search ability of NSGA-II. In this section, we compare four versions of NSGA-II: original NSGA-II with no discretization, NSGA-II-DD, NSGA-II-OD and NSGA-II-BD (i.e., NSGA-II with the both-space discretization).

### A. Algorithm

In NAGA-II-BD with the both-space discretization, NSGA-II is modified as follows. The SD-based decision space discretization of Kondo et al. [5] is applied to each solution just after they are generated. That is, decision variables are discretized before objective values are calculated. Then the objective values of each discretized solution are calculated. The calculated objective values are discretized by the above-mentioned simple 2-digit discretization method before non-dominated sorting. The basic framework of NSGA-II-BD with simultaneous discretization of both spaces is shown as in Algorithm 1.

| **Algorithm 1:** *NSGA-II-BD with the Both-space Discretization* |
|---|
| 1  Initialize $P_0$ |
| 2  *Discretize decision variables based on SD indicator* |
| 3  Evaluate $P_0$ |
| 4  **For** $t = 0$ : max generation **begin** |
| 5   Generate $P_t$ using tournament selection |
| 6   Generate offspring $Q_t$ using SBX and polynomial mutation |
| 7   *Discretize decision variables based on SD indicator* |
| 8   Evaluate $Q_t$ |
| 9   Combine $P_t$ and $Q_t$ into $R_t$ |
| 10  *Non-dominated sorting based on discretized objective function values* |
| 11  Generate $P_{t+1}$ from $R_t$ according to the result of non-dominated sorting and crowding distance sorting |
| 12 **end** |

### B. Experimental Results.

TABLE VII. AVERAGE IGD VALUES OBTAINED BY NSGA-II, NSGA-II-DD, NSGA-II-OD AND NSGA-II-BD ON THE LARGE-SCALE MANY-OBJECTIVE PROBLEMS WITH 1000 DECISION VARIABLES. THE BEST RESULT FOR EACH TEST PROBLEM IS HIGHLIGHTED BY THE BOLD BLUE FONT

| Problem | M | NSGA-II | NSGA-II-DD | NSGA-II-OD | NSGA-II-BD |
|---|---|---|---|---|---|
| DTLZ1 | 5 | 2.3189e+4 (9.00e+2) | 2.1591e+4 (9.49e+2) + | 2.1803e+4 (8.89e+2) + | **2.0967e+4 (1.01e+3) +** |
|  | 10 | 2.2464e+4 (1.64e+3) | 2.0440e+4 (1.59e+3) + | 1.9905e+4 (1.12e+3) + | **1.9763e+4 (1.10e+3) +** |
|  | 15 | 2.1389e+4 (1.72e+3) | 1.9915e+4 (1.45e+3) + | **1.9218e+4 (1.10e+3) +** | 1.9491e+4 (1.25e+3) + |
| DTLZ2 | 5 | 6.8200e+1 (2.75e+0) | 6.2799e+1 (2.21e+0) | 5.2614e+1 (2.05e+0) + | **5.0697e+1 (1.94e+0) +** |
|  | 10 | 8.4203e+1 (3.02e+0) | 8.1841e+1 (2.64e+0) + | 7.0448e+1 (2.10e+0) + | **7.0186e+1 (1.78e+0) +** |
|  | 15 | 8.1503e+1 (2.36e+0) | 8.0139e+1 (1.71e+0) + | 7.2821e+1 (1.83e+0) + | **7.2151e+1 (1.84e+0) +** |
| DTLZ3 | 5 | 9.3609e+4 (1.52e+3) | 8.9984e+4 (1.47e+3) + | 8.5260e+4 (1.32e+3) + | **8.2007e+4 (1.72e+3) +** |
|  | 10 | 1.0346e+5 (1.85e+3) | 1.0326e+5 (1.85e+3) = | 9.1816e+4 (1.57e+3) + | **9.1118e+4 (1.52e+3) +** |
|  | 15 | 1.0354e+5 (1.62e+3) | 1.0218e+5 (2.19e+3) + | **9.3129e+4 (1.31e+3) +** | 9.3908e+4 (1.97e+3) + |
| DTLZ4 | 5 | 6.9035e+1 (2.44e+0) | **3.5057e+1 (3.13e+0) +** | 4.6558e+1 (2.08e+0) + | 3.7750e+1 (4.55e+0) + |
|  | 10 | 8.1924e+1 (2.69e+0) | 7.1741e+1 (4.16e+0) + | 5.5545e+1 (1.45e+0) + | **5.0079e+1 (2.39e+0) +** |
|  | 15 | 8.1002e+1 (2.13e+0) | 7.6035e+1 (2.69e+0) + | 5.1841e+1 (1.40e+0) + | **4.8143e+1 (1.42e+0) +** |
| DTLZ5 | 5 | 7.3048e+1 (1.90e+0) | 7.0232e+1 (2.12e+0) + | 5.3793e+1 (1.97e+0) + | **5.1953e+1 (1.91e+0) +** |
|  | 10 | 7.9378e+1 (2.23e+0) | 7.9079e+1 (3.09e+0) = | **7.1202e+1 (1.78e+0) +** | 7.1924e+1 (1.54e+0) + |
|  | 15 | 7.9734e+1 (2.17e+0) | 7.8126e+1 (2.35e+0) + | **7.2233e+1 (1.81e+0) +** | 7.2418e+1 (1.33e+0) + |
| WFG1 | 5 | 2.0781e+0 (3.28e-2) | 2.4447e+0 (9.83e-3) - | **2.0214e+0 (2.14e-2) +** | 2.1688e+0 (3.94e-2) - |
|  | 10 | 3.2051e+0 (7.31e-2) | 3.3639e+0 (3.66e-2) - | **3.0559e+0 (4.67e-2) +** | 3.1640e+0 (3.39e-2) + |
|  | 15 | 4.1828e+0 (9.12e-2) | 4.3417e+0 (4.04e-2) - | **3.9664e+0 (3.97e-2) +** | 4.0722e+0 (3.86e-2) + |
| WFG2 | 5 | 1.2532e+0 (9.18e-2) | 1.0747e+0 (5.09e-2) + | 1.1810e+0 (7.92e-2) + | **1.0548e+0 (5.26e-2) +** |
|  | 10 | **2.5560e+0 (3.68e-1)** | 2.6361e+0 (3.67e-1) = | 2.9897e+0 (3.30e-1) - | 2.8549e+0 (4.73e-1) - |
|  | 15 | 1.9650e+0 (4.80e-1) | **1.8280e+0 (3.31e-1) =** | 3.6457e+0 (6.37e-1) - | 2.7414e+0 (4.39e-1) - |
| WFG3 | 5 | 1.2769e+0 (4.47e-2) | 1.2827e+0 (3.94e-2) = | 1.1576e+0 (3.43e-2) + | **1.1180e+0 (2.96e-2) +** |
|  | 10 | 2.6726e+0 (1.78e-1) | 2.6624e+0 (1.56e-1) = | 2.5095e+0 (1.44e-1) + | **2.4936e+0 (1.37e-1) +** |
|  | 15 | 4.1729e+0 (4.26e-1) | 4.2166e+0 (2.78e-1) = | 4.0056e+0 (3.76e-1) = | **3.9788e+0 (3.14e-1) +** |
| WFG4 | 5 | **1.3949e+0 (2.83e-2)** | 1.4043e+0 (2.45e-2) = | 1.4372e+0 (1.84e-2) - | 1.4710e+0 (3.16e-2) - |
|  | 10 | **4.9726e+0 (4.58e-2)** | 5.0115e+0 (4.96e-2) - | 5.0214e+0 (5.59e-2) - | 5.0567e+0 (5.73e-2) - |
|  | 15 | **8.9643e+0 (9.25e-2)** | 9.0926e+0 (8.77e-2) - | 9.0721e+0 (1.68e-1) - | 9.1201e+0 (9.64e-2) - |
| WFG5 | 5 | **1.5281e+0 (2.73e-2)** | 1.5358e+0 (2.62e-2) = | 1.5800e+0 (3.07e-2) - | 1.5636e+0 (2.45e-2) - |
|  | 10 | **5.0541e+0 (4.99e-2)** | 5.0711e+0 (4.31e-2) = | 5.0806e+0 (5.23e-2) - | 5.1008e+0 (5.07e-2) - |
|  | 15 | **8.9359e+0 (7.67e-2)** | 8.9789e+0 (8.72e-2) = | 8.9741e+0 (1.28e-1) = | 9.0080e+0 (8.88e-2) - |
| + / − / = |  |  | 14/5/11 | 21/7/2 | 21/9/0 |

Experimental results on the large-scale many-objective problems with 1000 decision variables are summarized in Table VII. The best result among the four algorithms is highlighted by the bold blue font for each problem (i.e., in each row). The statistical test is applied to the results by each of the three discretization strategies against the results by the original NSGA-II. The performance of NSGA-II is improved by the both-space discretization (NSGA-II-BD) for 21 out of the 30 problems. These results indicate the usefulness of the both-space discretization. Good results are also obtained by the objective space discretization (NSGA-II-OD). Whereas the decision space discretization (NSGA-II-DD) worked very well in Section III for the large-scale three-objective problems, it is not so effective on the large-scale many-objective problems.

Table VII clearly shows that the effect of the discretization depends on the problem. NSGA-II-BD works very well on DTLZ1-4 and WFG3. NSGA-II-OD shows the best results on DTLZ5 and WFG1. The usefulness of the discretization is unclear for WFG2. For WFG4 and WFG5, the best results are obtained by the original NSGA-II with no discretization. The discretization is not necessary or harmful for these problems. Since the usefulness of the discretization is problem dependent, it is an interesting future research topic to analyze the features of problems for which the discretization is useful. It is also an interesting future research topic to analyze the relation between an appropriate discretization strategy (i.e., one of DD, OD and BD) for each problem and the features of the problem.

### C. Further Investigation on DTLZ4

To further discuss the reason why the performance of NSGA-II can be improved by the discretization, we examine the obtained solution set of each discretization method for the

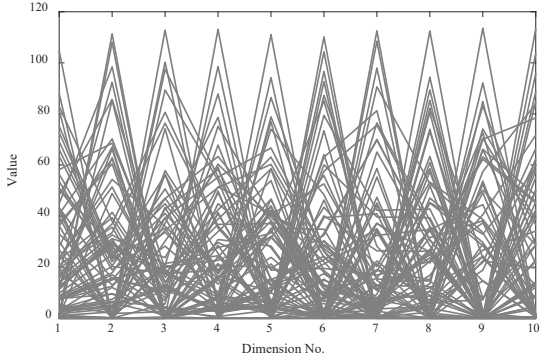

(a) NSGA-II with no discretization.

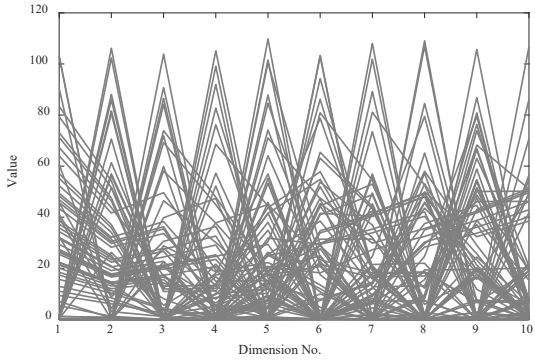

(b) NSGA-II-DD with the decision space discretization.

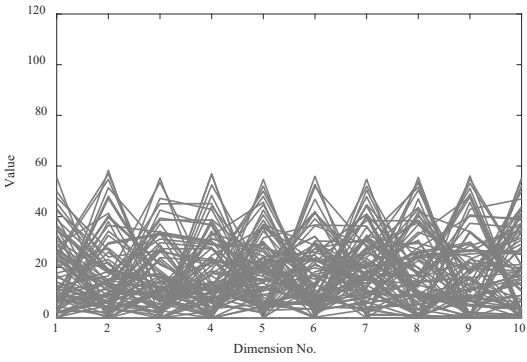

(c) NSGA-II-OD with the objective space discretization.

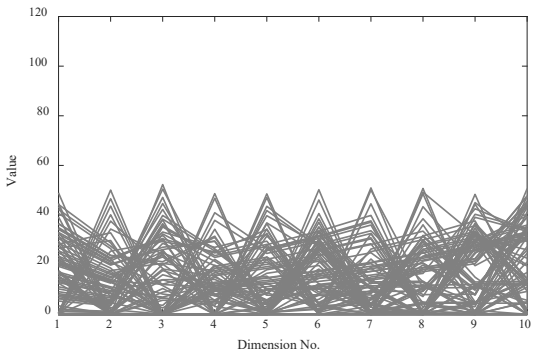

(d) NSGA-II-BD with the discretization in both decision and objective spaces.

**Fig. 5.** Obtained solutions by a single run of each algorithm on the large-scale 10-objective DTLZ4 problem with 1000 decision variables.

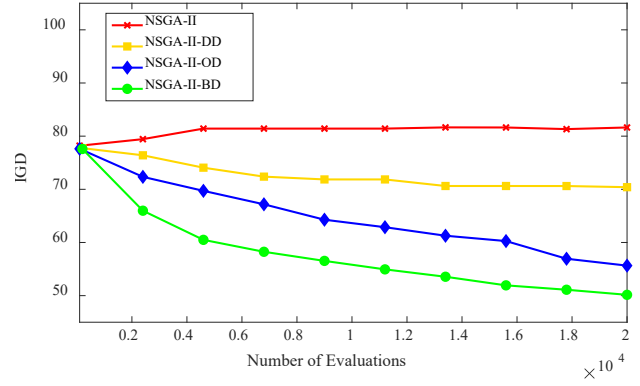

(a) IGD-based comparison of the four algorithms.

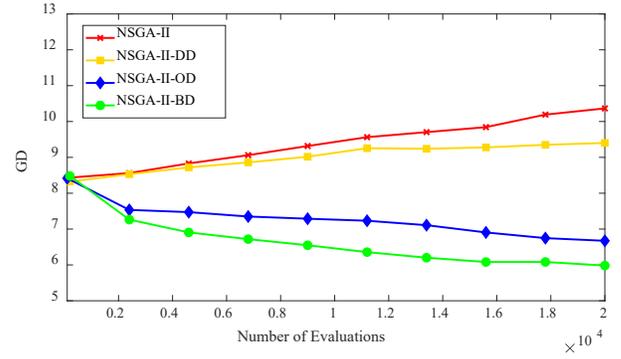

(b) GD-based comparison of the four algorithms.

**Fig. 6.** IGD and GD values at each generation of a single run (the same run as in Fig. 5) of each algorithm on the large-scale 10-objective DTLZ4 problem with 1000 decision variables.

10-objective DTLZ4 problem with 1000 decision variables. As shown in Table VII, the performance of NSGA-II is improved by each discretization strategy (i.e., DD, OD, BD). The best and the second-best results are obtained for this problem by NSGA-II-BD and NSGA-II-OD in Table VII, respectively. The obtained solution set by each algorithm is shown in Fig. 5. It suggests that the convergence ability of NSGA-II is significantly improved by the objective space discretization (i.e., OD). We can also observe a small positive effect of the decision space improvement (DD). As a result, the best result is obtained by BD (i.e., OD + DD).

For supporting these discussions, we show in Fig. 6 the IGD and GD values at each generation of the single run of each algorithm in Fig. 5. One interesting observation is that both the IGD and GD values increases throughout the execution of NSGA-II. This means that the initial population of NSGA-II is better than the final population. Since almost all solutions in each population are non-dominated, fitness evaluation is mainly based on crowding distance. Thus NSGA-II simply increases the diversity of solutions. As a result, the GD value continues to increase throughout the execution of NSGA-II (i.e., the population continues to move away from the Pareto front instead of approaching the Pareto front). The use of the decision space discretization (DD) slightly improves the convergence ability. However, the GD value continues to increase throughout the execution of NSGA-II-DD. The use of the objective space discretization (OD) clearly improves the convergence ability of NSGA-II. This is because the objective space discretization decreases the number of non-dominated solutions in each population as explained in Section IV. Since the decision space discretization (DD) has a small positive effect, the best results are obtained from NSGA-II-BD with the discretization on both decision and objective spaces (i.e., DD + OD). By comparing Fig. 6 (a) with Fig. 6 (b), we can see that the improvement in the overall performance measured by the IGD indicator is mainly due to the convergence improvement measured by the GD indicator.

## VI. CONCLUSION

In this paper, we examined the effects of the discretization of the decision and/or objective spaces on the performance of NSGA-II on large-scale multi-objective, many-objective and large-scale many-objective problems. The following is the general observations we obtained.

(i) The decision space discretization (DD) improved the search ability of NSGA-II for large-scale multi-objective problems with three objectives and 1000 decision variables.

(ii) The objective space discretization (OD) improved the search ability of NAGA-II for many-objective problems with 5-15 objectives and 9-19 decision variables.

(iii) The discretization of both decision and objective spaces (BD) improved the search ability of NSGA-II for large-scale many-objective problems with 5-15 objectives and 1000 decision variables. The best results were obtained by BD for those problems among the three strategies of discretization: DD, OD and BD.

Our first two observations extended the reported results in the previous literature (i.e., we showed the effect of DD on large-scale multi-objective problems rather than standard multi-objective problems in the literature [5] and the effect of OD on DTLZ and WFG problems rather than combinatorial problems in the literature [6], [7]). We also demonstrated the further improvement by combining the two discretization strategies: DD and OD (i.e., BD = DD + OD). Whereas the discretization of both decision and objective spaces (BD) improved the search ability of NSGA-II for many problems, BD also degraded NSGA-II for some problems (e.g., WFG4 and WFG5). The objective space discretization (OD) worked well on unmoral problems whereas it did not work well on multimodal problems. That is, the usefulness of each discretization strategy is problem dependent whereas the search ability of NSGA-II was improved by the three discretization strategies for many problems. An interesting future research topic is to analyze the relation between the effectiveness of each discretization strategy on each test problem and the characteristic features of the problem. The development of an adaptive method for the objective space discretization is also an interesting future research topic since only a simple objective space discretization method has been examined in the literature including this paper.